\documentclass[a4paper]{SIGEport}
\usepackage{graphicx}
\usepackage[caption=false, font=footnotesize]{subfig}
\usepackage{geometry}                                   
\usepackage{cite}
\begin{document}

\title{A Proposal for an Agentic AI Architecture to Support Multi-Domain Decision-Making in the Brazilian Armed Forces}

\author{Gioliano de Oliveira Braga$^1$, Sidnei Barbieri$^1$, Ágney Lopes Roth Ferraz$^1$, Wagner Comin Sonaglio$^1$,\\Henrique Curi de Miranda$^1$ e Lourenço Alves Pereira Jr.$^1$\\
	{\small $^1$Instituto Tecnológico de Aeronáutica (ITA), São José dos Campos/SP - Brasil}\\
\thanks{\setlength{\parindent}{0pt}\hspace{-1em}G. O. Braga, giolianobraga@ita.br; S. Barbieri, sidneisb@ita.br; Á. L. R. Ferraz, roth@ita.br; W. C. Sonaglio, sonaglio@ita.br; H. C. de Miranda, henriquecuri1@proton.me; L. A. Pereira Jr., ljr@ita.br.}}

\maketitle

\begin{abstract}
The growing complexity of multi-domain operational environments (land, aerospace, naval, cyber, and electromagnetic spectrum) has increased the volume and velocity of data reaching command-and-control (C2) centers, straining the observe-orient-decide-act (OODA) decision cycle. Artificial Intelligence (AI) systems currently employed in defense are, in general, reactive and isolated tools that still rely heavily on human operators to integrate information, assess scenarios, and formulate courses of action. This paper proposes a conceptual \textit{Agentic AI} architecture for AI systems that can plan, access data sources, execute tools, and act autonomously and audibly, aimed at supporting decision-making across the three Brazilian Armed Forces (Navy, Army, and Air Force). Four application fronts are discussed (decision support, situational analysis, feasibility studies, and countermeasure suggestion), as well as the data and sensor access requirements and the security and permission safeguards necessary for responsible employment across administrative, strategic, operational, and tactical contexts.
\end{abstract}


\section{Introduction}

The contemporary defense environment is characterized by growing integration across the land, naval, aerospace, cyber, and electromagnetic-spectrum domains. Modern joint operations produce a volume of data,  originating from radars, electro-optical sensors, electronic warfare systems, unmanned platforms, communication networks, and logistics and intelligence databases that far exceeds the capacity for manual analysis at a pace compatible with operational tempo (``OPTEMPO''). In this scenario, the observe-orient-decide-act (OODA) decision cycle becomes the critical factor of advantage: the Force that can turn raw data into informed decisions faster tends to retain operational initiative.

In recent years, the field of Artificial Intelligence (AI) has undergone a paradigm shift relevant to this problem. Until around 2022, systems based on Large Language Models (LLMs) were predominantly \textit{passive}: they received an input and returned a single output, with no capability to plan, use external tools, or execute multiple steps toward a goal. Since then, the concept of \textit{Agentic AI} has been consolidated: AI systems that operate in a continuous loop of reasoning, action, and observation, capable of querying databases, invoking Application Programming Interfaces (APIs), interacting with sensors and other computational tools, and adjusting their action plan based on the results obtained repeating this cycle autonomously until a goal is achieved or human intervention is required.

This paradigm has rapidly spread to the critical infrastructure and security domains. Recent architectures proposed for sixth-generation (6G) telecommunications networks, for example, already adopt Agentic AI frameworks based on multiple specialized agents: a real-time data triage agent, a threat classification agent supported by retrieval-augmented generation (RAG), and a response agent that operates directly on network control interfaces with the explicit goal of reducing security-incident response latency and enabling autonomous defense operations at scale~\cite{mobillm2025}. This type of architecture illustrates a principle that applies directly to the Brazilian military context: \textit{Agentic AI does not replace the human decision-maker, but drastically reduces the time between the detection of a relevant event and the presentation of well-founded courses of action to that decision-maker}.

At the same time, recent literature on AI agent security has shown that these systems introduce a new risk surface~\cite{safetyscale2025}: agents performing long-horizon tasks can fail in ways that are difficult to diagnose~\cite{horizon2026}; multi-agent systems that communicate through agent-to-agent protocols can be targeted by prompt injection attacks, inter-agent impersonation, and tool manipulation~\cite{a2asecbench2026,maltool2026}; and the very definition of ``security'' for an agent i.e., whether a given action represents legitimate behavior or a violation depends on context: who issued the instruction, what objective is being pursued, and whether the action actually serves that objective~\cite{formalizing2026}. Such findings are especially relevant for a defense environment, where the cost of an incorrect decision or an unauthorized action can be extremely high.

Given this scenario, the present work aims to propose a conceptual Agentic AI architecture for multi-domain decision support across the three Brazilian Armed Forces, the Brazilian Navy (\textit{Marinha do Brasil}, MB), the Brazilian Army (\textit{Ex\'ercito Brasileiro}, EB), and the Brazilian Air Force (\textit{For\c{c}a A\'erea Brasileira}, FAB). The proposal is not restricted to a single employment domain but seeks a reference model applicable to different decision levels (administrative, strategic, operational, and tactical) and warfare domains (land, air, naval, submarine, space, cyber, and electromagnetic).

The remainder of this paper is organized as follows. Section~\ref{sec:related} presents a brief discussion of related work in AI applied to defense and in Agentic AI. Section~\ref{sec:framing} presents the proposed framing for this paper's contribution. Section~\ref{sec:architecture} details the proposed conceptual architecture, organized into four application fronts. Section~\ref{sec:access} discusses the data and sensor access requirements, as well as the necessary security and permission safeguards. Section~\ref{sec:discussion} discusses limitations and risks. Finally, Section~\ref{sec:conclusion} presents the conclusions and proposes directions for future work.

\begin{figure*}[htbp!]
    \centering
    \includegraphics[width=1\linewidth]{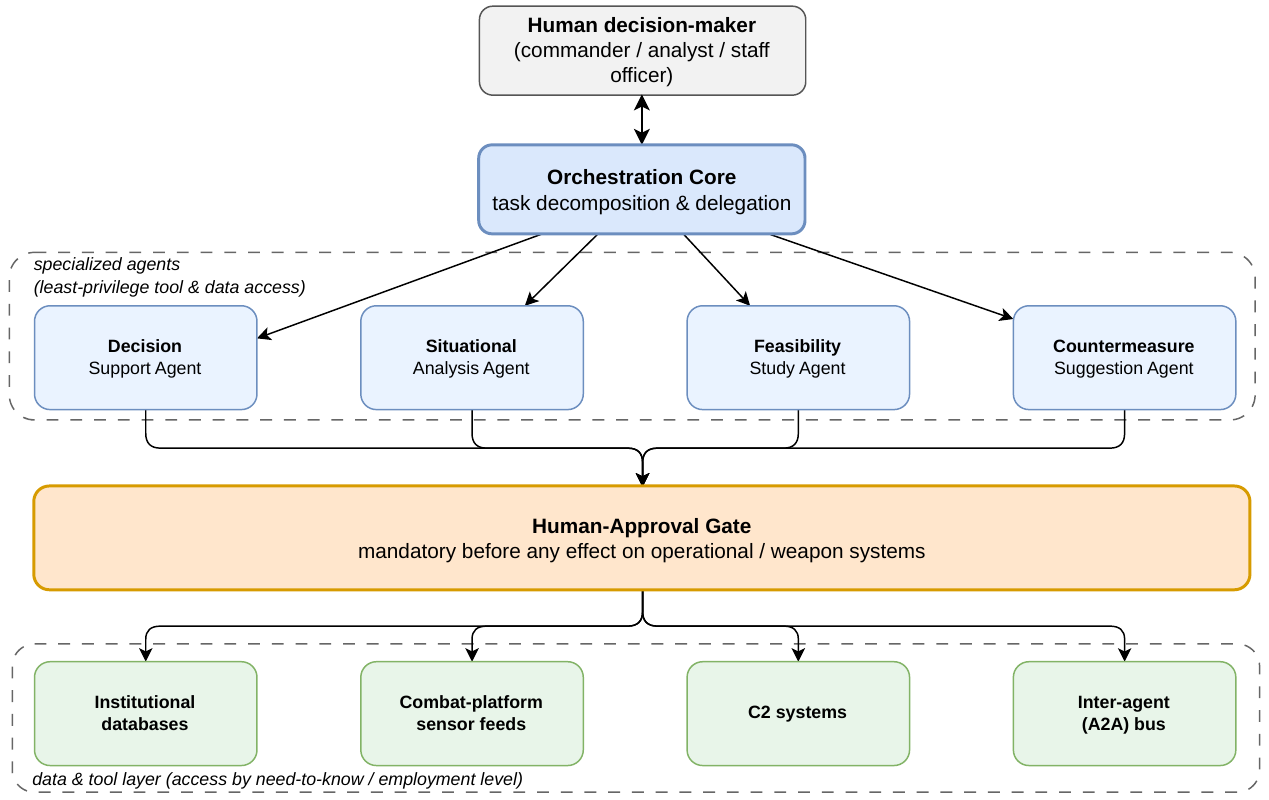}
    \caption{Reference architecture for the proposed Agentic AI decision-support layer. The orchestration core decomposes a request from the human decision-maker and delegates subtasks to specialized agents, each restricted to a subset of the data and tool layer. Recommendations that would affect operational or weapon systems must first pass a mandatory human-approval gate, preserving meaningful human control over the highest-impact decisions.}
    \label{fig:architecture}
\end{figure*}

\section{Related Work}
\label{sec:related}

The use of AI in defense applications is not new: multi-criteria decision support systems, pattern detection in radar signals, and classification of Synthetic Aperture Radar (SAR) images are well-established examples of \textit{passive} AI applied to operational problems~\cite{safetyscale2025}. These systems, however, typically operate as isolated modules: a classifier identifies a target, a sensor-fusion system estimates its position, and it falls to the human operator to manually integrate these outputs into a situational picture and decide on a course of action.

Recent literature on Agentic AI documents the transition from this isolated paradigm to integrated architectures. Frameworks that combine multiple specialized agents each responsible for one stage of the decision process (detection, classification, recommendation, and execution), have been proposed for domains such as telecommunications network security, where integration with trusted knowledge bases (e.g., technical specifications from regulatory bodies) is cited as an essential element for ensuring the reliability of the agent's decisions~\cite{mobillm2025}  although such retrieval-augmented pipelines themselves introduce knowledge-corruption risks that must be accounted for~\cite{poisonedrag2025}.

Another relevant line of research evaluates agent performance on long-horizon tasks, i.e., tasks requiring extended, interdependent sequences of actions. Recent studies show that LLM-based agents, while performing satisfactorily on short tasks, tend to degrade significantly as the task horizon increases, with failures attributable to categories such as planning errors, catastrophic forgetting of context information, and accumulation of errors over the action history~\cite{horizon2026}.
This finding has direct implications for the military employment of agents: defense operations frequently involve long sequences of interdependent decisions (for example, from tracking an emerging threat to recommending a countermeasure), so the agent's robustness over time must be treated as a design requirement rather than an emergent property.

Finally, a third line of research deals specifically with the security of agents and multi-agent systems. Recent work proposes frameworks for formalizing the notion of ``contextual security'' in LLM-based agents~\cite{formalizing2026}, decomposing it into properties such as task alignment (the agent pursues authorized objectives), action alignment (each individual action serves those objectives), source authorization (the agent executes only commands from authenticated sources), and data isolation (information flows respect privilege boundaries). Other work demonstrates, through benchmarks specific to agent-to-agent protocols, that LLM-based multi-agent architectures are vulnerable to inter-agent impersonation attacks, manipulation of task orchestration, and the injection of malicious instructions into artifacts exchanged between agents, with attack success rates that can reach 100\% in implementations without adequate safeguards~\cite{a2asecbench2026}. These results reinforce the need for any Agentic AI proposal for defense to treat the security of the agent itself as a core part of the architecture, rather than as an afterthought.

\section{Framing of the Contribution}
\label{sec:framing}

To situate this paper's contribution, it is useful to make explicit the adopted framing, i.e., how the problem, the gap identified in the literature, and the proposed contribution relate to one another.

\textbf{Problem (broad context).} The three Brazilian Armed Forces increasingly operate in a multi-domain battlespace, in which information from land, naval, air, space, cyber, and electromagnetic-spectrum sensors must be correlated in near-real time to support decisions at multiple levels, from administrative resource-management decisions to tactical decisions on the employment of assets in combat.

\textbf{Gap.} Most AI applications currently discussed in the Brazilian defense context focus on perception tasks (detection, classification, tracking) and operate as isolated systems, without an \textit{integrating-reasoning} component capable of combining these perceptions, autonomously querying multiple data sources, and proposing, in a traceable and auditable manner, decision options to the commander or operator. Moreover, the discussion of how to integrate such capabilities securely while respecting the chain of command, information classification levels, and each Force's principles of employment remains underexplored in the national literature.

\textbf{Contribution.} This paper proposes (i) a reference architecture for Agentic AI applicable to all three Forces, organized into four application fronts decision support, situational analysis, feasibility studies, and countermeasure suggestion; (ii) a characterization of the data, system, and sensor access requirements necessary for such agents to operate; and (iii) a set of security and permission-control safeguards, organized by level of employment (administrative, strategic, operational, and tactical), grounded in recent literature on AI agent security.

Thus, this paper does not propose a new machine-learning technique, but rather an \textit{architectural and governance reference model} for the responsible adoption of Agentic AI in the national defense context, a contribution of a conceptual and systems-engineering nature, aligned with SIGE's thematic areas of Command and Control, Cyber Defense, and Multi-Agent Systems.

\section{Proposed Architecture}
\label{sec:architecture}

\begin{figure*}[htbp!]
\centering
\includegraphics[width=1\textwidth]{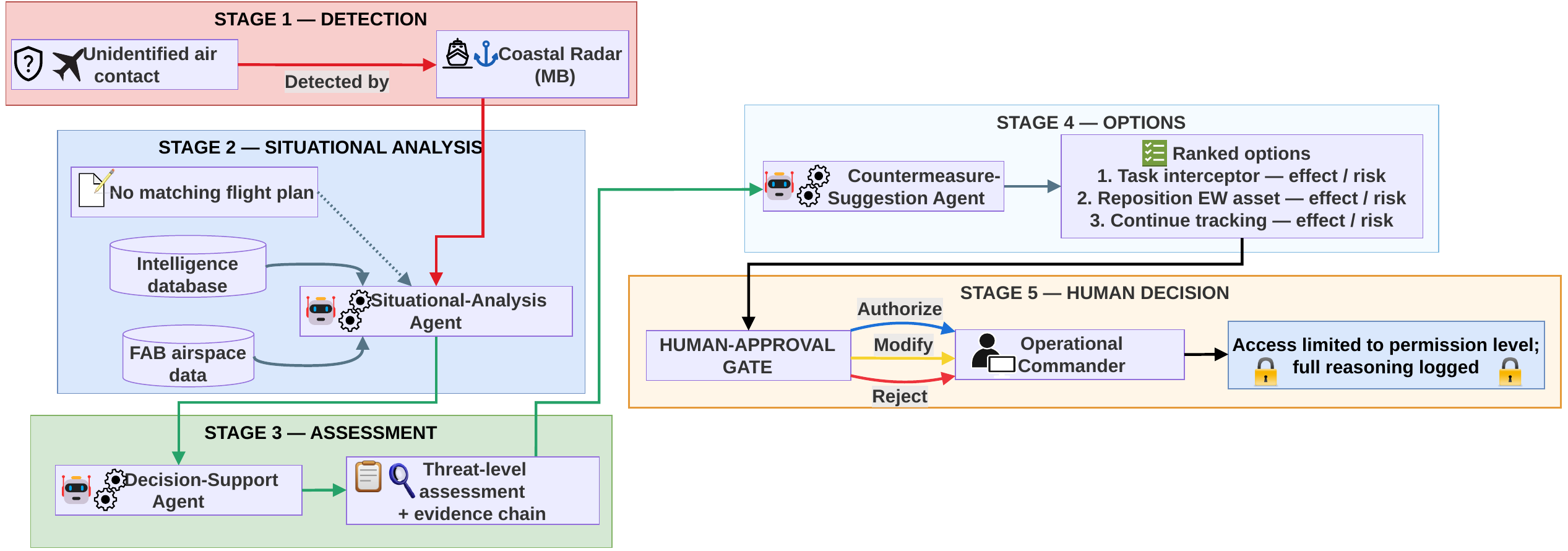}
\caption{End-to-end workflow of the proposed architecture in a joint
air-defense vignette, showing multi-source correlation, evidence-backed
recommendation, and mandatory human authorization before any effect.}
\label{fig:scenario}
\end{figure*}

The proposed architecture is organized around an \textit{orchestration core}, a coordinating agent (a commander, an analyst, or a staff officer) that receives the user's request, decomposes it into subtasks, and delegates them to specialized agents. Figure~\ref{fig:architecture} presents this reference architecture, in which every recommendation that could affect operational or weapon systems must traverse a human-approval gate before taking effect. Each specialized agent has access to a restricted set of tools and data sources, according to its role; Figure~\ref{fig:scenario} illustrates the resulting workflow in a joint-operation example. The following subsections describe the four identified application fronts.

\subsection{Decision Support}

In this role, the agent serves as an analytical assistant in the decision-making process, without supplanting the authority of the human decision-maker. Example tasks include:

\begin{itemize}
    \item \textbf{Threat-level assessment:} the agent correlates data from multiple sources (e.g., radar detections, electromagnetic signal intercepts, and intelligence reports) to estimate the probability and severity of a threat, presenting the operator not only with the estimate but also with the chain of evidence supporting it.
    \item \textbf{Route recommendation:} in transport, resupply, or troop-movement operations, the agent can query geospatial databases, weather forecasts, and risk-area reports to propose alternative routes, making explicit the advantages and risks of each option.
    \item \textbf{Prioritization of targets or tasks:} in scenarios with multiple concurrent events, the agent can assist in prioritization based on criteria previously defined by the command (e.g., criticality, window of opportunity, availability of assets).
\end{itemize}

In all of these cases, a fundamental design requirement is \textit{explainability}~\cite{trustgen2025}: the agent's recommendation must be accompanied by a justification traceable to the data sources consulted, so that the human decision-maker can validate or reject the suggestion based on their own judgment and the rules of engagement in force.

\subsection{Situational Analysis}

This front addresses the continuous construction and updating of a situational awareness picture. The agent can be employed to:

\begin{itemize}
    \item \textbf{Capability comparison:} consolidate, from intelligence databases and catalogs of own assets, a structured comparison between the capabilities of an adversary force and the capabilities available to the friendly force in a given area of operations, highlighting relevant asymmetries.
    \item \textbf{Identification of limitations:} point out sensor coverage gaps, logistical constraints (fuel, ammunition, spare parts), or time windows during which a given capability will be unavailable.
    \item \textbf{Continuous monitoring:} maintain an updated summary of the current situation, proactively flagging significant changes (e.g., the appearance of a new contact or a change in the readiness state of an adversary unit).
\end{itemize}

\subsection{Feasibility Studies}

This front applies mainly to administrative decisions and medium- to long-term planning, such as the acquisition of new systems, the allocation of resources between units, or the choice between modernization alternatives. In these cases, the agent can:

\begin{itemize}
    \item Consolidate information on cost, schedule, technical performance, and risk associated with each alternative, based on technical documents, vendor proposals, and historical data from similar projects;
    \item Perform comparative analyses (e.g., cost-benefit analysis, sensitivity analysis with respect to schedule or budget variations);
    \item Present the decision-maker with a structured synthesis of the alternatives, including their respective strengths, weaknesses, and underlying assumptions, while keeping the final decision within the human and institutional sphere.
\end{itemize}

\subsection{Countermeasure Suggestion}

This front demands the greatest caution, as it is the one most directly related to the operational and tactical domains. Here, the agent can assist in generating countermeasure options across different warfare domains, such as:

\begin{itemize}
    \item \textbf{Land domain:} suggesting unit repositioning or formation adjustments in response to a detected threat;
    \item \textbf{Air and space domain:} recommending evasive maneuvers, flight-route adjustments, or reallocation of space-surveillance assets;
    \item \textbf{Naval and submarine domain:} suggesting changes in course, speed, or depth in response to the detection of a surface or subsurface threat;
    \item \textbf{Cyber domain:} identifying attack signatures and suggesting containment actions (network-segment isolation, blocking of source addresses);
    \item \textbf{Electromagnetic-spectrum domain:} recommending electronic-warfare techniques (e.g., adjusting electronic countermeasure parameters) in response to the detection of a new emitter.
\end{itemize}

It is essential to emphasize that, in this front in particular, the \textbf{proposed architecture does not include the autonomous execution of kinetic or offensive actions}. The agent's role is limited to \textit{suggesting} and \textit{prioritizing} options for human evaluation, preserving the principle of meaningful human control over the use of force, in accordance with Brazilian military doctrine and applicable law~\cite{end2020}.

\section{Access Requirements and Security Safeguards}
\label{sec:access}

The usefulness of an AI agent is directly related to its ability to access relevant, up-to-date data. At the same time, recent literature on agent security shows that every new access granted to an agent (be it to a database, a software tool, or a communication channel with another agent) also represents a new attack surface~\cite{autosafe2025,safetyscale2025}. This section discusses, in an integrated manner, the necessary accesses and their corresponding safeguards as shown in Figure~\ref{fig:architecture}.

\subsection{Required Types of Access}

\begin{itemize}
    \item \textbf{Integration with institutional databases:} logistics, maintenance, human resources, and intelligence systems, required mainly for the feasibility study and situational analysis fronts.
    \item \textbf{Access to combat-platform sensor data:} radars, electro-optical/infrared systems, electronic warfare systems, and navigation systems, required for situational analysis and countermeasure suggestion in near-real time.
    \item \textbf{Access to Command and Control (C2) systems:} so that the agent's recommendations can be contextualized within the current battlespace picture and, where applicable, presented directly within the interfaces already used by operators.
    \item \textbf{Inter-agent communication (multi-agent systems):} in joint operations, agents from different Forces or command levels may need to exchange information, following agent-to-agent communication protocols.
\end{itemize}

\subsection{Permission Levels by Type of Employment}

Consistent with military hierarchy and with the principles of need-to-know and least privilege, and aligned with national command-and-control doctrine~\cite{md31m03}, it is proposed that the agent's access and autonomy be differentiated according to the type of employment~\cite{ragrbac2026}:

\begin{itemize}
    \item \textbf{Administrative employment:} greater autonomy for querying and consolidating non-sensitive data (e.g., logistics, financial, and maintenance data). The agent's actions are limited to generating reports and analyses; no automated action directly affects operational systems.
    \item \textbf{Strategic employment:} access to consolidated intelligence and higher-level planning data, but with autonomy restricted to producing analyses and scenarios; decisions remain entirely within the scope of general officers and the Ministry of Defense.
    \item \textbf{Operational employment:} access to C2 data at the theater-of-operations level, with autonomy to correlate information and propose courses of action, always subject to approval by the operational commander.
    \item \textbf{Tactical employment:} potentially real-time access to sensor data from a specific platform or unit, with autonomy limited to immediate alerts and suggestions; any action that produces an effect on weapon systems remains under direct human control.
\end{itemize}

\subsection{Recommended Technical Safeguards}

Based on recent literature on AI agent security, it is recommended that the architecture incorporate, from the design stage, the following mechanisms:

\begin{itemize}
    \item \textbf{Data and privilege isolation:} ensure that information flows originating from external or less trustworthy sources cannot alter the agent's behavior with respect to internal, authenticated sources, mitigating prompt injection attacks~\cite{privsep2025,critical2025};
    \item \textbf{Source authorization:} every command or instruction processed by the agent must be attributable to an authenticated source within the chain of command, with full logging for later audit~\cite{formalizing2026};
    \item \textbf{Task and action alignment:} mechanisms for continuously verifying that the agent's actions remain consistent with the originally authorized objective, flagging deviations (task drift) for human review~\cite{taskdrift2025};
    \item \textbf{Validation of external tools and agents:} in multi-agent architectures, verification of the identity and integrity of partner agents before exchanging information, in order to mitigate inter-agent impersonation risks~\cite{a2asecbench2026,maltool2026,toolhijacker2025};
    \item \textbf{Human supervision at critical decision points:} explicit definition, for each application front and level of employment, of the workflow points at which human approval is mandatory before any of the agent's recommendations produces a practical effect~\cite{agentwatcher2026}.
\end{itemize}

\section{Implementation and Evaluation Considerations}
\label{sec:impl}

Although the architecture is model-agnostic, a prototype can be built from existing components. The orchestration core and specialized agents can be implemented using agent-orchestration frameworks such as LangGraph, Microsoft AutoGen, or CrewAI, which provide native support for planning, tool invocation, and multi-agent coordination. To meet sovereignty and information-classification constraints, the underlying models can be open, nationally hostable LLMs (e.g., the Llama or Mistral families) served on institutional infrastructure~\cite{metasecalign2025}. Situational grounding can be provided by a retrieval-augmented generation (RAG) layer atop a vetted knowledge base of doctrine and technical specifications, mirroring the standards-grounded design adopted in the telecommunications domain~\cite{mobillm2025}. Tool and inter-agent connectivity can follow emerging standard protocols, whose adoption also delimits the attack surface that the safeguards of Section~\ref{sec:access} are designed to protect~\cite{a2asecbench2026}.

Because the contribution is architectural, its validation should rest on an explicit evaluation protocol rather than on a single accuracy figure. We propose assessing a future prototype, in a simulated joint exercise, along five dimensions: (i) \textit{decision-cycle gain}, the reduction in OODA time-to-option relative to the current human-only process; (ii) \textit{recommendation quality}, measured as precision and recall against expert-defined courses of action; (iii) \textit{traceability}, the fraction of recommendations delivered with a complete evidence chain to their sources; (iv) \textit{safety}, quantified as the attack-success rate under prompt injection and the task-drift detection rate on agentic benchmarks adapted to the defense setting~\cite{taskdrift2025}; and (v) \textit{human-oversight cost}, comprising the human-override rate and the time-to-approval at the human-approval gate. Robustness should additionally be reported as a function of task horizon, since agent reliability degrades on longer action sequences~\cite{horizon2026}.

\section{Discussion: Limitations and Risks}
\label{sec:discussion}

The proposal presented in this paper is conceptual in nature, and its practical validation would require, among other aspects, the implementation of a prototype in a controlled environment (e.g., a simulated exercise), the detailed definition of communication protocols between agents and the three Forces' legacy systems, and a formal assessment of compliance with current legislation and doctrine governing the employment of autonomous systems in defense.

Additionally, it should be considered that LLM-based agents may exhibit performance degradation on long-horizon tasks, with error accumulation over extended sequences of actions~\cite{horizon2026}. In a military context, this reinforces the need for intermediate verification mechanisms and for limiting the horizon of autonomy granted to the agent without supervision, especially in the countermeasure-suggestion and tactical-level decision-support fronts. Reliability concerns are further compounded in defense-oriented workflows such as cyber threat intelligence, where heterogeneous and rapidly evolving evidence can induce domain-specific failure modes~\cite{cti2025}.

Finally, it is worth noting that the adoption of any solution based on large language models in a defense environment raises additional questions regarding technological sovereignty, for example, the preference for open, auditable models that can be hosted on national or institutional infrastructure rather than external services~\cite{metasecalign2025}, which fall outside the technical scope of this paper, but are central to its future implementation.

\section{Conclusion}
\label{sec:conclusion}

This paper proposed a framing and a conceptual architecture for employing Agentic AI as a decision-support layer for multi-domain operations across the three Brazilian Armed Forces. The central contribution consists of organizing, in an integrated manner, four application fronts-decision support, situational analysis, feasibility studies, and countermeasure suggestion-around an orchestration core of specialized agents, and of characterizing, for each level of employment (administrative, strategic, operational, and tactical), the corresponding data and sensor access requirements and security safeguards.

Unlike proposals that treat AI as a set of isolated perception tools, the perspective adopted here emphasizes the agent's role as an \textit{integrator of information and reasoning}, whose function is to reduce the time between the detection of relevant events and the presentation of well-founded options to the human decision-maker, preserving, in all cases, meaningful human control over the highest-impact decisions.

As future work, we intend to: (i) implement a prototype of the proposed architecture and evaluate it, in a simulated joint exercise, against the metrics defined in Section~\ref{sec:impl}; (ii) detail a case study in a specific domain, such as naval patrol or integrated aerospace defense; and (iii) address its integration with the three Forces' legacy command-and-control systems and its compliance with national doctrine.

\bibliographystyle{IEEEtran}
\bibliography{RefSIGE}

\end{document}